\documentclass{article}
\usepackage{spconf,amsmath,graphicx}
\ninept 
\usepackage[utf8]{inputenc}
\usepackage{psfrag,epsfig,graphics}
\usepackage{amsmath,amsthm,amssymb,multirow}
\usepackage{mathbbol}
\usepackage{amssymb}             

\DeclareSymbolFontAlphabet{\amsmathbb}{AMSb}%
\usepackage{mathabx} 

\usepackage{booktabs}
\usepackage{graphicx}

\usepackage[noadjust]{cite}
\usepackage{multirow}
\usepackage[lined,linesnumbered,ruled]{algorithm2e}

\usepackage{color}  
\def\cred{\textcolor{red}}

\def\cblue{\textcolor{blue}}

\newcommand{\cb}[1]{{\boldsymbol{#1}}}
\newcommand{\cp}[1]{\ifmmode {\mathcal{#1}}\else ${\mathcal{#1}}$\fi}

\newcommand{\bA}{\boldsymbol{A}}

\newcommand{\bI}{\boldsymbol{I}}

\newcommand{\bM}{\boldsymbol{M}}

\newcommand{\bR}{\boldsymbol{R}}

\newcommand{\bX}{\boldsymbol{X}}

\newcommand{\bZ}{\boldsymbol{Z}}

\newcommand{\bm}{\boldsymbol{m}}

\newcommand{\be}{\boldsymbol{e}}

\newcommand{\br}{\boldsymbol{r}}

\newcommand{\bx}{\boldsymbol{x}}
\newcommand{\bz}{\boldsymbol{z}}

\newcommand{\bPsi}{\boldsymbol{\Psi}}

\newcommand{\balpha}{\boldsymbol{\alpha}}

\newcommand{\bbM}{\amsmathbb{M}}
\newcommand{\bbX}{\amsmathbb{X}}
\newcommand{\bbR}{\amsmathbb{R}}
\newcommand{\bbPsi}{\mathbb{\Psi}}
\newcommand{\bbPhi}{\mathbb{\Phi}}

\title{IMPROVED HYPERSPECTRAL UNMIXING WITH ENDMEMBER VARIABILITY PARAMETRIZED USING AN INTERPOLATED SCALING TENSOR}
\name{Ricardo~Augusto~Borsoi, Tales Imbiriba, Jos\'e~Carlos~Moreira~Bermudez
\thanks{This work has been supported by the National Council for Scientific and Technological Development (CNPq).}
\thanks{R. A. Borsoi, T. Imbiriba and J.C.M. Bermudez are with the Department of Electrical Engineering, Federal University of Santa Catarina, Florian\'opolis, SC, Brazil. e-mail: raborsoi@gmail.com; talesim@gmail.com; j.bermudez@ieee.org.}%
}
\address{Federal University of Santa Catarina, Florianópolis, SC, Brazil}
\allowdisplaybreaks
\begin{document}
%


\maketitle
\begin{abstract}
Endmember (EM) variability has an important impact on the performance of hyperspectral image (HI) analysis algorithms. Recently, extended linear mixing models have been proposed to account for EM variability in the spectral unmixing (SU) problem. The direct use of these models has led to severely ill-posed optimization problems. Different regularization strategies have been considered to deal with this issue, but none so far has consistently exploited the information provided by the existence of multiple pure pixels often present in HIs. 
In this work, we propose to break the SU problem into a sequence of two problems. First, we use pure pixel information to estimate an interpolated tensor of scaling factors representing spectral variability. This is done by considering the spectral variability to be a smooth function over the HI and confining the energy of the scaling tensor to a low-rank structure. Afterwards, we solve a matrix-factorization problem to estimate the fractional abundances using the variability scaling factors estimated in the previous step, what leads to a significantly more well-posed problem. Simulations with synthetic and real data attest the effectiveness of the proposed strategy.
\end{abstract}
\section{Introduction}
\label{sec:intro}

Spectral Unmixing (SU) is part of a number of algorithms that retrieve vital information from hyperspectral images (HIs) in many applications~\cite{Keshava:2002p5667}.
SU aims at extracting the spectral signatures of materials present in the HI of a scene, as well as the proportion in which they contribute to each HI pixel.  
Many parametric models have been proposed to describe the interaction between light and the target surface~\cite{Keshava:2002p5667, Dobigeon-2014-ID322}.
The simplest of such models is the \textit{Linear Mixing Model} (LMM), which considers that the observed reflectance of an HI pixel is obtained from a convex combination of the spectral signatures of pure materials. This model imposes a convex geometry to the SU problem, where all HI pixels are confined to a simplex whose vertices are the reflectances of pure materials, usually termed endmembers (EMs). The linearity and convexity of the LMM model lead to an interpretation of its coefficients as the relative abundances of each pure material in the HI.
Nevertheless, some characteristics of practical HIs cannot be modeled by the standard LMM, such as nonlinearities~\cite{Dobigeon-2014-ID322, Imbiriba2016_tip, Imbiriba2017_bs_tip, Imbiriba2014} or variations of the EMs along the image~\cite{Thouvenin_IEEE_TSP_2016_PLMM, drumetz2016blindUnmixingELMM, imbiriba2018glmm}.  More sophisticated models are required when such nonidealities have important impact on the HI.

EM variability can result, for instance, from environmental conditions, illumination, atmospheric or temporal changes~\cite{Zare-2014-ID324-variabilityReview}. Its occurrence may incur  the propagation of significant estimation errors 
throughout the unmixing process~\cite{Thouvenin_IEEE_TSP_2016_PLMM}. Most of the methods
proposed so far to deal with spectral variability can be classified in three major groups: endmembers as sets, endmembers as statistical distributions and, more recently, methods that incorporate the variability in the mixing model, often using physically motivated concepts~\cite{drumetz2016variabilityReviewRecent}. The method proposed in this work combines elements from the first and third groups. Specifically, we adopt a modified mixing model over a very specific variational set. 

In the first group, EM variability is addressed by considering different types of variational sets, among which spectral bundles are prominent~\cite{Zare-2014-ID324-variabilityReview}. One nice property of these methods is that the bundles can be directly extracted from the observed HI.
%
Few works expand this idea by adopting pixel-dependent EMs obtained using some kind of spatial interpolation of the original EM signatures~\cite{maselli2001estimatingSpatiallyVariableEndmembers, li2016geostatisticalEndmemberInterpolation,johnson2012spatialInterpolationEndmembers}. These interpolated EMs sets are then used to unmix the data using the standard LMM. This type of approach often profits from \textit{a priori} information about EM spectra (pure pixels) and their position in the image. Interpolation strategies include linear regression~\cite{maselli2001estimatingSpatiallyVariableEndmembers, zhang2015neighbourhoodSpecificEndmemberSignatureGeneration, li2017classificationAndEndmemberInterpolation} and kriging~\cite{li2016geostatisticalEndmemberInterpolation, johnson2012spatialInterpolationEndmembers}.
However, these strategies lack flexibility to adapt to variations in the endmember signatures that lie beyond the descriptive capability of the spectral bundles.

In the third group, different extensions of the LMM have been proposed to cope with spectral variability~\cite{Thouvenin_IEEE_TSP_2016_PLMM, drumetz2016blindUnmixingELMM, imbiriba2018glmm}. The models differ with respect to the type of variability they represent and their physical motivation. The Perturbed LMM model (PLMM)~\cite{Thouvenin_IEEE_TSP_2016_PLMM} introduces an additive perturbation to the EM matrix. It allows the modeling of arbitrary spectral variations, but lacks physical motivation. The Extended LMM (ELMM)~\cite{drumetz2016blindUnmixingELMM} and its generalization, the Generalized LMM (GLMM)~\cite{imbiriba2018glmm}, are physically motivated, but differ in their ability to model arbitrary variability.
%
%
The ELMM performs well if spectral variability is due mainly to illumination variations, but it lacks flexibility when the EMs are subject to more complex spectral distortions. The GLMM is able to model arbitrary EM variability by considering band-dependent multiplicative factors, thus linking the amount of spectral variability to the magnitude of the EM reflectance in each band. This effect is consistent with empirical observations, leading to a more physically reasonable model.
%
%
One drawback of extended LMMs is that they lead to ill-posed optimization problems. Hence, regularization strategies must be devised to yield meaningful results~\cite{drumetz2017relationshipsBilinearELMM, Borsoi_2018_Fusion}. 
%
Different regularization strategies have been recently proposed.  For instance, \cite{Borsoi2017_multiscale} and~\cite{Borsoi_multiscaleVar_2018}  proposed a data-dependent multiscale strategy to exploit the spatial uniformity existing in HIs. The approaches in~\cite{qian2017tensorNMFunmixing, imbiriba2018ULTRA} assume that most of the energy of HIs is confined to a low-dimensional structure within the image space, and exploit low-rank tensor decomposition techniques to impose regularity to the solution. 







In this work we propose a hybrid approach by leveraging pure pixel information in the context of parametric endmember models. Specifically, we propose to break the SU problem into two sequential problems. We first use pure pixel information to estimate an interpolated tensor of spectral variability scaling factors by assuming that its energy is confined to a low-rank structure. Afterwards, we solve a matrix-factorization problem to estimate the fractional abundances. This is done by using the variability scaling factors estimated in the previous step in the form of a regularization term, which leads to a significantly more well-posed problem. We adopt the GLMM to parametrically model spectral variability, as it combines flexibility and physical motivation.

This paper is organized as follows. Section~\ref{sec:lmms} briefly reviews the LMM and its GLMM extended version. Section~\ref{sec:prop} introduces the proposed formulation and the corresponding SU algorithm. The performance of the proposed method is compared with competing algorithms in Section~\ref{sec:Simulations}. Finally, the conclusions are presented in Section~\ref{sec:conclusions}.

\vspace{-0.3cm}
\section{Linear Mixing Models}\label{sec:lmms}
The Linear Mixing Model (LMM)~\cite{Keshava:2002p5667} assumes that a given a pixel $\br_n = [r_{n,1},\,\ldots, \,r_{n,L} ]^\top$, with $L$ bands, is represented as
\begin{align}
 &\br_n = \bM \balpha_n + \be_n, \label{eq:LMM}
\quad \text{subject to }\,\cb{1}^\top\balpha_n = 1 \text{ and } \balpha_n \succeq \cb{0} 
\end{align}
where $\bM \in \amsmathbb{R}^{L\times R}$ is a matrix whose columns are the $R$ endmember spectral signatures $\bm_k$, $\balpha_n$ is the abundance vector, $\be_n$ is an additive white Gaussian noise (WGN) and $\succeq$ is the entrywise $\geq$ operator. 
The LMM assumes that the endmember spectra are fixed for all pixels $\br_n$, $n=1,\ldots,N$, in the HI. This assumption jeopardizes the accuracy of estimated abundances in many circumstances due to the spectral variability existing in a typical scene.

To mitigate this issue, the GLMM~\cite{imbiriba2018glmm} models arbitrary EM variability by considering band-dependent multiplicative factors, which links the amount of spectral variability to the magnitude of the EM reflectance in each band.


The GLMM introduces a scaling matrix $\cb{\Psi}_n\in{\amsmathbb{R}^{L\times R}}$ with entries $[\cb{\Psi}_n]_{\ell,k} = \psi_{n_{\ell,k}}\geq 0$, which acts individually on each wavelength. The model represents the $n$-th HI pixel as
\begin{equation} \label{eq:model_glmm}
\br_n = (\bM\odot\cb{\Psi}_n)\balpha_n + \be_n
\end{equation}
where $\odot$ stands for the Hadamard product.

\section{Proposed Unmixing Algorithm}
\label{sec:prop}

Given a parametric extended linear mixing model $f$, and an HI $\bR = [\br_1,\ldots,\br_N]$, the SU problem can be generally cast as the determination of $\bA$ and $\bbM$ that minimize a risk functional of the form
\begin{equation} \label{eq:general_unmxing_cf}
 J(\bA,\cb{\bbM})=  \frac{1}{2}\|\bR - f(\bA,\cb{\bbM})\|_F^2 + \mathcal{R}(\bA) + \mathcal{R}(\bbM)
\end{equation}
where $\bbM$ is a 3-D tensor obtained by stacking all pixel-dependent endmember matrices $\bM_n$, such that $[\bbM]_{n,:,:} = \bM_{n}$, $\bA$ is the abundance matrix, and $\mathcal{R}(\bA)$ and $\mathcal{R}(\bbM)$ are regularization terms to improve the problem conditioning.
Although parametric EM models such as the ELMM and GLMM have enough flexibility to adequately represent the variability in the endmember spectra, the resulting unmixing problem is highly ill-posed.
%
%
%
This problem is accentuated by the fact that previous formulations such as~\cite{drumetz2017relationshipsBilinearELMM, imbiriba2018glmm} only considered a single reference, or average endmember matrix, as a priori information (or initialization), while estimating the variability maps and endmembers for all pixels in the scene. 
Although good performance has been reported in~\cite{drumetz2017relationshipsBilinearELMM, imbiriba2018glmm}, the correct estimation of the parametric models for complex scenes remains a challenge.

A characteristic of many scenes, which remains unexplored in this context, is the existence of multiple pure pixels in an observed HI. This information can be leveraged to help in the estimation of the parametric endmember model, introducing more information into the problem and reducing its ill-posedness.

We propose to break the unmixing problem into a sequence of two problems: 
\begin{itemize}
 \item[(i)] Using pure pixel information extracted from the HI by standard EM extraction algorithms and assuming smooth spectral variability throughout the HI, estimate an interpolated tensor of variability factors $\bbPsi$ whose energy is assumed to be confined to a low-rank structure.
 \item[(ii)] Using the estimated tensor as a prior information in a standard matrix-factorization problem, solve it to estimate the endmember and abundance matrices.
\end{itemize}

\vspace{-0.2cm}
\subsection{Estimating the Tensor of Variability Factors}

The objective of this first problem is to estimate the scaling matrices $\bPsi_n$, $n=1,\ldots, N$, in \eqref{eq:model_glmm}. To this end, we assume:
\begin{itemize}
 \item[a)] the availability of a reference endmember matrix $\bM_0$, which can be obtained using any endmember extraction method;
 \item[b)] the knowledge of the set $\mathcal{N}_{\mathcal{P},k}$ of locations $(n_1,n_2)$ of pure pixels for the $k$-th endmember, for all $k=1,\dots,R$; \footnote{Pure pixels are here defined here as a set of pixels whose spectral distance relative to the reference EMs in $\bM_0$ is less than a specified threshold.}
 \item[c)] that the EM variability changes smoothly in the HI, so that each endmember variability function has most of its energy confined to a low-dimensional structure in the image space. 
\end{itemize}


Consider the 4-D tensor $\bbPsi\in\amsmathbb{R}^{N_1\times N2\times L\times R}$, $N_1\times N_2=N$, of variability weights with entries $[\bbPsi]_{n_1,n_2,\ell,k} = \psi_{n_1,n_2,\ell,k}$, corresponding to the scaling factor to be applied to the $\ell$-th band of the $k$-th endmember in the pixel at position $(n_1,n_2)$ of the HI. $\bbPsi$ generalizes the GLMM scaling factors.  We propose to estimate the variability tensor $\bbPsi$ by solving the following optimization problem: 
\begin{align} \label{eq_OptimizationProblem}
    J_{\bbPsi}(\bbPsi,\bbPhi) & {}={}
    \|\bbPsi - \bbPhi\|_F^2  + \varepsilon \|\bbPsi - \mathbb{1}\|_F^2 
    \\[-0.1cm] & \hspace{0.0cm}
    + \lambda_{\bbPsi} \sum_{k=1}^R \sum_{(n_1,n_2)\in\mathcal{N}_{\mathcal{P},k}}
    \|\widetilde{\br}_{n_1,n_2}-\bm_{0,k} \odot [\bbPsi]_{n_1,n_2,:,k} \|^2 \nonumber
    \\ & 
    \text{subject to } \,\,\,  \operatorname{rank}(\bbPhi) = r  \nonumber
\end{align}
In \eqref{eq_OptimizationProblem}, $\bbPhi$ is a low-rank tensor which imposes a low-rank constraint on the variability maps to enforce the smooth variation. $\mathbb{1}$ is an $N_1\times N_2\times L\times R$ tensor of ones, so that $\|\bbPsi -\mathbb{1}\|^2_F$ is a regularization term that enforces the variability to be small, and whose importance is controlled by the parameter $\varepsilon>0$. $\widetilde{\br}_{n_1,n_2}$ represents the pixel at position $(n_1,n_2)$, $\bm_{0,k}$ is the $k$-th column of $\bM_0$ with elements $m^0_{\ell,k}$, and the third term enforces the similarity between pure pixels at known positions $(n_1,n_2)$ and their corresponding endmembers and variability maps. Its strictness is controlled by $\lambda_\bbPsi > 0$. The associated optimization problem is given by
\begin{align}\label{eq:PsiPhiProblem}
    \min_{\bbPsi,\bbPhi} \,\,\, J_{\bbPsi}(\bbPsi,\bbPhi)
\end{align}

This problem is non-convex and generally NP-hard due to the rank constraint. We propose to solve it using the alternating optimization approach described in Algorithm~\ref{alg:psi_opt}, as follows.

\begin{algorithm} 
\scriptsize
\SetKwInOut{Input}{Input}
\SetKwInOut{Output}{Output}
\caption{Algorithm for solving~\eqref{eq:PsiPhiProblem}~\label{alg:psi_opt}}
\Input{$\bR$, $\lambda_{\bbPsi}$, $r$, $\{\mathcal{N}_{\mathcal{P},k}\}_k$.}
\Output{$\bbPsi^*$.}
Set $i=0$ and $\bbPhi^{(0)}=\mathbb{1}$ \;
\While{stopping criterion is not satisfied}{
$i=i+1$ \;
$\bbPsi^{(i)} = \underset{\bbPsi}{\arg\min} \,\,\,\,  {J}_{\bbPsi}(\bbPsi,\bbPhi^{(i-1)})$ \;
$\bbPhi^{(i)} = \underset{\bbPhi}{\arg\min} \,\,\,\,  {J}_{\bbPsi}(\bbPsi^{(i)},\bbPhi)$ \;
}
\KwRet $\widehat{\bbPsi}=\bbPsi^{(i)}$  \;
\end{algorithm}

\subsubsection{Optimizing with respect to $\bbPsi$}
To solve problem~\eqref{eq:PsiPhiProblem} with respect to $\bbPsi$ we consider $\bbPhi$ fixed. Thus, the optimization problem with respect to $\bbPsi$ is given by
\begin{align}
    \min_{\bbPsi}  &  \,\,\, 
    \|\bbPsi - \bbPhi\|_F^2  + \varepsilon \|\bbPsi - \mathbb{1}\|_F^2 
    \nonumber \\[-0.15cm] &
    + \lambda_{\bbPsi} \sum_{k=1}^R \sum_{(n_1,n_2)\in\mathcal{N}_{\mathcal{P},k}}
    \|\widetilde{\br}_{n_1,n_2}-\bm_{0,k} \odot [\bbPsi]_{n_1,n_2,:,k} \|^2
\end{align}
By taking the derivative with respect to each position $[\bbPsi]_{n_1,n_2,\ell,k}$ of $\bbPsi$ and setting it equal to zero, we end up with the following solution:
\begin{align}  
    & [\widehat{\bbPsi}]_{n_1,n_2,\ell,k} 
    \\ \nonumber & \hspace{4ex} =
    \begin{cases}
    \displaystyle{\frac{[\bbPhi]_{n_1,n_2,\ell,k} + \varepsilon}{1+\varepsilon}}, & (n_1,n_2)\notin \mathcal{N}_{\mathcal{P},k}\\
    \displaystyle{\frac{[\bbPhi]_{n_1,n_2,\ell,k} + \varepsilon + \lambda_{\bbPsi}\, m^0_{\ell,k}\,\widetilde{r}_{\ell,n_1,n_2}}{1+\varepsilon+\lambda_{\bbPsi}\, (m^0_{\ell,k})^2}}, & (n_1,n_2)\in \mathcal{N}_{\mathcal{P},k}
    \end{cases}
    %
\end{align}
for $\ell=1,\ldots,L$, $k=1,\ldots,R$ and $n_i=1,\ldots,N_i$, \, $i=1,2$.

\subsubsection{Optimizing with respect to $\bbPhi$}

The optimization problem with respect to $\bbPhi$ is given by
\begin{align}
    & \min_{\bbPhi}\,\,\, 
    \|\bbPsi - \bbPhi\|_F^2
    \quad \text{subject to } \,\,\, \operatorname{rank}(\bbPhi) = r \label{eq:PhiProblem}
\end{align}
where we write tensor $\bbPhi$ as 
\begin{equation}
 	\bbPhi = \sum_{i=1}^{r} 		
    \xi_{i}\,\bz^{(1)}_i\circ\bz^{(2)}_{i}\circ\bz^{(3)}_{i}\circ\bz^{(4)}_{i}.
 \label{eq:Phi_cpd}
\end{equation}
where $\circ$ denotes the outer product and $r$ is a small number since we assume that most of the energy of $\bbPsi$ lies in a low-rank structure. Using~\eqref{eq:Phi_cpd} in~\eqref{eq:PhiProblem} leads to the following optimization problem:
\begin{align}
 \Big(\widehat{\cb{\Xi}}\,\,,&\,\,\widehat{\bZ}^{(1)}\!,\widehat{\bZ}^{(2)}\!,\widehat{\bZ}^{(3)}\!,\widehat{\bZ}^{(4)}\Big) = \label{eq:Phi_als}
 \\[-0.15cm] &
 \mathop{\arg\min}_{\cb{\Xi},\bZ^{(1)},\bZ^{(2)},\bZ^{(3)},\bZ^{(4)}}\Big\|\bbPsi - \sum_{i=1}^{r} \xi_{i}\bz_i^{(1)}\circ\bz_i^{(2)}\circ\bz_i^{(3)}\circ\bz_i^{(4)}\Big\|^2_F \nonumber
\end{align} 
where $\cb{\Xi} = \text{Diag}_4\big(\xi_{1}, \ldots, \xi_{r}\big)$ is an order-4 diagonal tensor with $[\cb{\Xi}]_{i,i,i,i}=\xi_i$.
Problem~\eqref{eq:Phi_als} can be solved using an alternating least-squares strategy~\cite{sidiropoulos2017tensor}.
The solution $\widehat{\bbPhi}$ is then obtained using the full multilinear product~\cite{imbiriba2018ULTRA}:
\begin{equation}
 \widehat{\bbPhi} = \big\ldbrack \widehat{\cb{\Xi}} \,\,;\, \widehat{\bZ}^{(1)}; \widehat{\bZ}^{(2)};\widehat{\bZ}^{(3)} \big\rdbrack.
\end{equation}


\subsection{The Spectral Unmixing Problem}
Given a reference EM matrix $\bM_0$ and the EM scaling factors $\bPsi_n = \widehat{\bbPsi}_{n,:,:}$ for each pixel (with each $n$ corresponding to a single pair $(n_1,n_2)$), the SU problem can be formulated as the minimization of the risk functional~\eqref{eq:general_unmxing_cf} with respect to~$\bA$ and~$\bbM$, using an appropriate EM model and regularization terms. We propose to minimize the following regularized cost functional:
\begin{align} \label{eq:glmm_cost_func}
    J(\bA,\cb{\bbM}) {}={} & \frac{1}{2} \!\sum_{n=1}^{N} \!\big(\|\br_n-\bM_n\balpha_n\|^2
    \!+\! \lambda_M\|\bM_n-\bM_0\odot\bPsi_n\|^2_{F} \big) 
    \nonumber \\[-0.1cm] &
    + \lambda_A \big( \|\cp{H}_h(\bA)\|_{2,1} + \|\cp{H}_v(\bA)\|_{2,1}\big) 
    \\ &
    \text{subject to } \,\,\, \bA\succeq0, \,\, \bA^\top \cb{1} = \cb{1}, \,\, \cb{\bbM}\succeq0
    \nonumber 
\end{align}
The last term in~\eqref{eq:glmm_cost_func} promotes spatial regularity in the abundance maps. $\mathcal{H}_h$ and $\mathcal{H}_v$ are linear operators that compute the horizontal and vertical gradients of a bidimensional signal, acting separately for each material of $\bA$, and $\|\bX\|_{2,1}=\sum_{n=1}^N\|\bx_n\|_2$ is the $\mathcal{L}_{2,1}$ norm.

%
Cost function in~\eqref{eq:glmm_cost_func} is non-smooth and non-convex with respect to both $\bA$ and $\bbM$, but convex with respect to each of them. Following the approaches used in~\cite{drumetz2016blindUnmixingELMM,imbiriba2018glmm}, we search for a locally optimal solution by alternately minimizing~\eqref{eq:glmm_cost_func} with respect to each variable, as described in Algorithm~\ref{alg:global_opt}.

\begin{algorithm} 
\scriptsize
\SetKwInOut{Input}{Input}
\SetKwInOut{Output}{Output}
\caption{Global algorithm for solving \eqref{eq:glmm_cost_func}~\label{alg:global_opt}}
\Input{$\bR$, $\lambda_M$, $\lambda_A$, $\lambda_{\bbPsi}$, $\bA^{(0)}$ and $\bM_0$.}
\Output{$\,\widehat{\!\bA}$, $\widehat{\bbM}$ and $\widehat{\bbPsi}$.}
Estimate $\bbPsi$ using Algorithm~\ref{alg:psi_opt} \;
Set $i=0$ \;
\While{stopping criterion is not satisfied}{
$i=i+1$ \;
$\bbM^{(i)} = \underset{\bbM}{\arg\min} \,\,\,\,  {J}(\bA^{(i-1)},\bbM,\bbPsi)$ \;
$\bA^{(i)} = \underset{\bA}{\arg\min} \,\,\,\,  {J}(\bA,\bbM^{(i)},\bbPsi)$ \;
}
\KwRet $\,\widehat{\!\bA}=\bA^{(i)}$,~ $\widehat{\bbM}=\bbM^{(i)}$,~ $\widehat{\bbPsi}=\bbPsi$  \;
\end{algorithm}

\subsubsection{Optimization with respect to $\bbM$}
Considering only the terms in~\eqref{eq:glmm_cost_func} that depend on $\bbM$, the optimization problem for the EM tensor becomes
\begin{align}
    & \mathop{\min}_{\bbM \succeq \cb{0}} \,\, \frac{1}{2} \!\sum_{n=1}^N \!\big(\|\br_n-\bM_n\balpha_n\|^2
    \!+\! \lambda_M\|\bM_n -\bM_0\odot\bPsi_n\|^2_F\big).
\label{eq:M_problem}
\end{align}
Problem~\eqref{eq:M_problem} can be solved individually for each pixel $\br_n$. Relaxing the nonnegativity constraint on $\bbM$, the solution is given by
\begin{equation}
\widehat{\!\bM}_n = (\br_n\balpha^\top+ \lambda_M\bM_0\odot\bPsi_n)(\balpha_n\balpha_n^\top + \lambda_M\bI_R)^{-1}
\end{equation}
An approximate solution to the original constrained problem~\eqref{eq:M_problem} can then be obtained by projecting $\,\widehat{\!\bM}_n$ onto the nonnegative orthant $\bbR_+^{L\times R}$ by attributing zero to any negative entries~\cite{drumetz2016blindUnmixingELMM,imbiriba2018glmm}.

\subsubsection{Optimization with respect to $\bA$}
Considering only the terms in~\eqref{eq:glmm_cost_func} that depend on $\bA$, the optimization problem for the abundance matrix becomes
\begin{align} \label{eq:A_problem}
    & \mathop{\min}_{\bA} \,\, \frac{1}{2} \! \sum_{n=1}^{N} \|\br_n - \bM_n\balpha_n\|^2
    \! + \lambda_A \big( \|\cp{H}_h(\bA)\|_{2,1} + \|\cp{H}_v(\bA)\|_{2,1}\big) 
    \nonumber 
    \\[-0.1cm] &
    \,\,\, \text{subject to } \,\,\, \bA\succeq0, \,\, \bA^\top \cb{1} = \cb{1}
\end{align}
This problem is non-smooth and not separable with respect to the different pixels in the HI. Nevertheless, problem~\eqref{eq:A_problem} can be solved efficiently using the Alternating Direction Method of the Multipliers (ADMM)~\cite{Boyd_admm_2011}. The procedure is described in detail in~\cite{drumetz2016blindUnmixingELMM}.

\section{Experimental Results} \label{sec:Simulations}
We now illustrate the performance of the proposed method through simulations with both synthetic and real data. We compare the proposed method with the fully constrained least squares (FCLS), the ELMM~\cite{drumetz2016blindUnmixingELMM}, the PLMM~\cite{Thouvenin_IEEE_TSP_2016_PLMM} and the GLMM~\cite{imbiriba2018glmm}. In all experiments, the reference EM matrix $\bM_0$ was extracted from the observed HI using the VCA algorithm~\cite{Nascimento2005}. The sets $\mathcal{N}_{\mathcal{P},k}$ of pure pixel locations were estimated by selecting the pixels with the smallest spectral angle relative to the reference EMs in $\bM_0$. The performances were evaluated using the Root Means Squared Error (RMSE) between the estimated abundance maps ($\text{RMSE}_{\bA}$), between the EM matrices ($\text{RMSE}_{\bbM}$) and between the reconstructed images ($\text{RMSE}_{\bR}$). The RMSE between two generic tensors is defined as
\begin{equation}
\text{RMSE}_{\bbX} = \sqrt{\textstyle{\frac{1}{N_{\bbX}}}\|\bbX- \left.\bbX\right.^*\|^2_F} 
\end{equation}
where $N_{\bbX}$ denotes the number of elements in the tensor $\bbX$.

We consider also the Spectral Angle Mapper (SAM) to evaluate the estimated EM tensor
\begin{equation}
{
  \text{SAM}_{\bbM} = \frac{1}{N}\sum_{n=1}^{N}\sum_{k=1}^{R}\arccos\bigg(\frac{\bm_{k,n}^\top\bm_{k,n}^*}{\|\bm_{k,n}\|\|\bm_{k,n}^*\|}\bigg).
}
\end{equation}



\subsection{Synthetic data}
A synthetic dataset was created from spatially correlated abundance maps containing a significant amount of pure pixels. The HI was constructed using reference EMs extracted from the USGS Spectral Library~\cite{clark2003imaging}. Spectral variability was introduced following the GLMM model in Eq.~\eqref{eq:model_glmm} to generate pixel-dependent EM signatures with spatial and spectral correlation imposed on $\bPsi_n$ using a 3-D Gaussian filter. Then, WGN was added to yield a 30dB SNR.

We selected the optimal parameters for each algorithm by performing a grid search using the ranges of parameters suggested by the authors in the original publications. 
For PLMM we used $\gamma=1$ and searched for $\alpha$ and $\beta$ in the range $\{0.1,\, 25\}$ and $\{10^{-9},\, 10^{-3}\}$, respectively. For ELMM, GLMM and the proposed method, we selected the parameters in the following ranges: $\lambda_{S},\,\lambda_M \in \{0.01, 50\}$, $\lambda_{A} \in \{0.001,\, 0.1\}$, and $\lambda_\psi,\,\lambda_{\bbPsi} \in \{10^{-4},\,10^3\}$. We also used $\varepsilon=10^{-5}$ and $r=10$ fixed in Algorithm~\ref{alg:psi_opt}.  Finally, the number of pure pixels extracted from the image was $500$, $100$ and $10$ for the first, second and third EMs, respectively.

The results are shown in Table~\ref{tab:results_synthData}. It can be seen that the proposed method yields a significantly smaller $\text{RMSE}_{\bA}$ than the other algorithms. It also yields the second-best results $\text{RMSE}_{\bbM}$ and $\text{SAM}_{\bbM}$, both very close to the best results obtained by the GLMM. 
The reconstruction errors are smaller for PLMM, GLMM and ELMM, which is expected since these methods have a larger number of degrees of freedom, while the proposed method estimates the variability scaling factors a priori. However, smaller reconstruction errors do not necessarily imply better abundance estimation, as evidenced by the $\text{RMSE}_{\bA}$ results. The execution time of the proposed method is $1.95\times \text{Time}_{\text{GLMM}}$, a modest increase in complexity when compared to the benefits in terms of estimation accuracy.

\begin{table} 
\footnotesize
\caption{Simulations with synthetic and real data.}
\centering
\renewcommand{\arraystretch}{1}
\scriptsize
\begin{tabular}{lcccc|c}
\bottomrule
\multicolumn{5}{c}{Synthetic HI}  &\multicolumn{1}{|c}{Houston HI}\\
\toprule\midrule
& $\text{RMSE}_{\bA}$ & $\text{RMSE}_{\bbM}$ &$\text{SAM}_{\bbM}$ & $\text{RMSE}_{\bR}$ & $\text{RMSE}_{\bR}$ \\ \midrule
FCLS & 0.0416 & -- & -- & 0.0212  &  0.0478 \\
PLMM & 0.0353 & 0.0220 & 0.0232  & \cred{0.0117}  & 0.0360 \\
ELMM & 0.0373 & 0.0135 & 0.0174  & 0.0121  & \cblue{0.0033} \\
GLMM & \cblue{0.0343} & \cred{0.0128} & \cred{0.0163} & \cblue{0.0118}   &  \cred{0.0003} \\
Proposed & \cred{0.0233} & \cblue{0.0130} & \cblue{0.0164}  & 0.0123   & 0.0085 \\
\bottomrule\toprule
\end{tabular}
\label{tab:results_synthData}
\end{table}
%
\begin{figure}
\centering
\includegraphics[height= 5.9cm, width=0.4\textwidth]{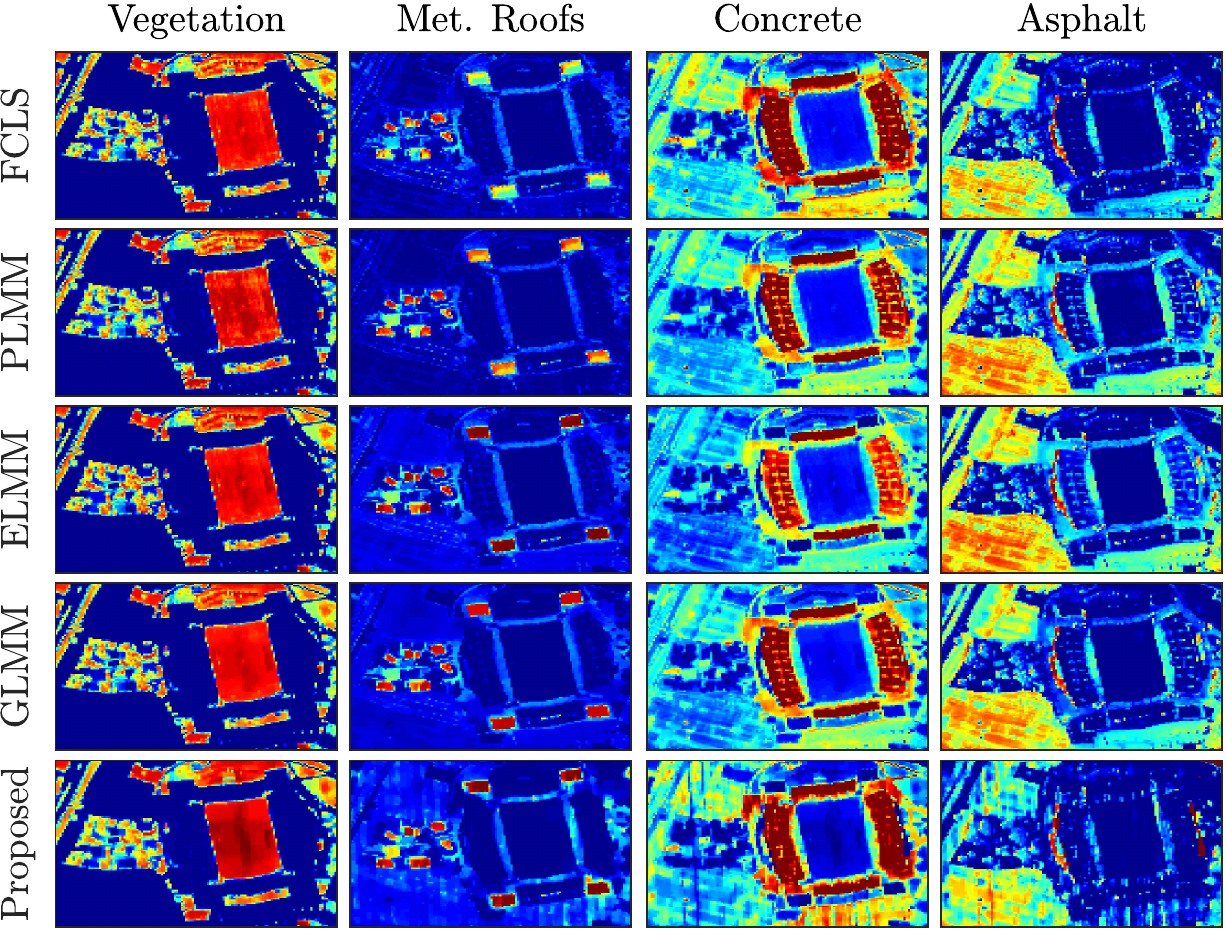}
\vspace{-2ex}
\caption{Abundance maps of the Houston dataset for all tested algorithms where the abundance values are represented by colors ranging from blue ($\alpha_k = 0$) to red ($\alpha_k = 1$).}\label{fig:ab_maps_houston}
\end{figure}

\subsection{Real data}
For simulations with real data we considered the Houston dataset, which is known to have four EMs~\cite{drumetz2016blindUnmixingELMM, imbiriba2018glmm}. The parameters for the proposed method were set as $\lambda_M=0.1$, $\lambda_A=0.01$, $\lambda_{\bbPsi}=10^3$, $\varepsilon=10^{-5}$ and $r=10$. For the other algorithms we used the same parameters~\cite{imbiriba2018glmm}. The sets $\mathcal{N}_{\mathcal{P},k}$ were constructed by extracting all pixels in the HI with an angle smaller than $\cos^{-1}(0.01)$ to the reference EMs. The reconstructed abundance maps for all tested algorithms are shown in Fig.~\ref{fig:ab_maps_houston}, while the reconstruction errors are presented in Table~\ref{tab:results_synthData}.

It can be seen that the proposed method provides abundance estimation that is accurate and comparable with the results obtained using ELMM or GLMM. Furthermore, the proposed method results in stronger concrete and vegetation components in the stadium stands and in the field, respectively, when compared to the other algorithms. As in the synthetic data example, the results in Table~\ref{tab:results_synthData} show smaller $\text{RMSE}_{\bR}$ values for GLMM and ELMM. Again, this is not directly related to better abundance estimation. The proposed method demanded a computational time equal to $2.77\times\text{Time}_{\text{GLMM}}$.


\vspace{-0.3cm}
\section{Conclusions}\label{sec:conclusions}
We have proposed a new spectral unmixing algorithm accounting for spectral variability. Modeling the spectral variability as a smooth function over the image, pure pixel information was leveraged from the HI to estimate a tensor of EM scaling factors prior to unmixing. The use of this predetermined tensor has led to a much more well-posed SU problem when compared to current algorithms, in which the scaling factors are blindly estimated from the image. Simulations using synthetic and real data indicate that the proposed method can lead to significant improvements in abundance estimation accuracy.


\bibliographystyle{IEEEbib}
\bibliography{strings,references}

\begin{thebibliography}{10}

\bibitem{Keshava:2002p5667}
N.~Keshava and J.~F. Mustard,
\newblock ``Spectral unmixing,''
\newblock {\em IEEE Signal Processing Magazine}, vol. 19, no. 1, pp. 44--57,
  2002.

\bibitem{Dobigeon-2014-ID322}
N.~Dobigeon, J.-Y. Tourneret, C.~Richard, J.~C.~M. Bermudez, S.~McLaughlin, and
  A.~O. Hero,
\newblock ``Nonlinear unmixing of hyperspectral images: Models and
  algorithms,''
\newblock {\em IEEE Signal Processing Magazine}, vol. 31, no. 1, pp. 82--94,
  Jan 2014.

\bibitem{Imbiriba2016_tip}
T.~Imbiriba, J.~C.~M. Bermudez, C.~Richard, and J.-Y. Tourneret,
\newblock ``Nonparametric detection of nonlinearly mixed pixels and endmember
  estimation in hyperspectral images,''
\newblock {\em IEEE Transactions on Image Processing}, vol. 25, no. 3, pp.
  1136--1151, March 2016.

\bibitem{Imbiriba2017_bs_tip}
T.~Imbiriba, J.~C.~M. Bermudez, and C.~Richard,
\newblock ``Band selection for nonlinear unmixing of hyperspectral images as a
  maximal clique problem,''
\newblock {\em IEEE Transactions on Image Processing}, vol. 26, no. 5, pp.
  2179--2191, May 2017.

\bibitem{Imbiriba2014}
T.~Imbiriba, J.~C.~M. Bermudez, J.-Y. Tourneret, and C.~Richard,
\newblock ``Detection of nonlinear mixtures using gaussian processes:
  Application to hyperspectral imaging,''
\newblock in {\em ICASSP, IEEE International Conference on Acoustics, Speech
  and Signal Processing}, Jan 2014, pp. 7949--7953.

\bibitem{Thouvenin_IEEE_TSP_2016_PLMM}
P.-A. Thouvenin, N.~Dobigeon, and J.-Y. Tourneret,
\newblock ``Hyperspectral unmixing with spectral variability using a perturbed
  linear mixing model,''
\newblock {\em IEEE Trans. Signal Processing}, vol. 64, no. 2, pp. 525--538,
  Feb. 2016.

\bibitem{drumetz2016blindUnmixingELMM}
L.~Drumetz, M.-A. Veganzones, S.~Henrot, R.~Phlypo, J.~Chanussot, and
  C.~Jutten,
\newblock ``Blind hyperspectral unmixing using an extended linear mixing model
  to address spectral variability,''
\newblock {\em IEEE Transactions on Image Processing}, vol. 25, no. 8, pp.
  3890--3905, 2016.

\bibitem{imbiriba2018glmm}
T.~Imbiriba, R.~A. Borsoi, and J.~C.~M. Bermudez,
\newblock ``Generalized linear mixing model accounting for endmember
  variability,''
\newblock in {\em Acoustics, Speech and Signal Processing (ICASSP), 2018 IEEE
  International Conference on}. IEEE, 2018, pp. 1862--1866.

\bibitem{Zare-2014-ID324-variabilityReview}
A.~Zare and K.~C. Ho,
\newblock ``Endmember variability in hyperspectral analysis: Addressing
  spectral variability during spectral unmixing,''
\newblock {\em Signal Processing Magazine, IEEE}, vol. 31, pp. 95--104, January
  2014.

\bibitem{drumetz2016variabilityReviewRecent}
Lucas Drumetz, Jocelyn Chanussot, and Christian Jutten,
\newblock ``Variability of the endmembers in spectral unmixing: recent
  advances,''
\newblock in {\em 8th IEEE Workshop on Hyperspectral Image and Signal
  Processing: Evolution in Remote Sensing}, Los Angeles, USA, 2016.

\bibitem{maselli2001estimatingSpatiallyVariableEndmembers}
Fabio Maselli,
\newblock ``Definition of spatially variable spectral endmembers by locally
  calibrated multivariate regression analyses,''
\newblock {\em Remote Sensing of Environment}, vol. 75, no. 1, pp. 29--38,
  2001.

\bibitem{li2016geostatisticalEndmemberInterpolation}
Wenliang Li and Changshan Wu,
\newblock ``A geostatistical temporal mixture analysis approach to address
  endmember variability for estimating regional impervious surface
  distributions,''
\newblock {\em GIScience \& Remote Sensing}, vol. 53, no. 1, pp. 102--121,
  2016.

\bibitem{johnson2012spatialInterpolationEndmembers}
Brian Johnson, Ryutaro Tateishi, and Toshiyuki Kobayashi,
\newblock ``Remote sensing of fractional green vegetation cover using
  spatially-interpolated endmembers,''
\newblock {\em Remote Sensing}, vol. 4, no. 9, pp. 2619--2634, 2012.

\bibitem{zhang2015neighbourhoodSpecificEndmemberSignatureGeneration}
Zhang Zhang, Chong Liu, Jiancheng Luo, Zhanfeng Shen, and Zhenfeng Shao,
\newblock ``Applying spectral mixture analysis for large-scale sub-pixel
  impervious cover estimation based on neighbourhood-specific endmember
  signature generation,''
\newblock {\em Remote Sensing Letters}, vol. 6, no. 1, pp. 1--10, 2015.

\bibitem{li2017classificationAndEndmemberInterpolation}
Wenliang Li and Changshan Wu,
\newblock ``A geographic information-assisted temporal mixture analysis for
  addressing the issue of endmember class and endmember spectra variability,''
\newblock {\em Sensors}, vol. 17, no. 3, pp. 624, 2017.

\bibitem{drumetz2017relationshipsBilinearELMM}
Lucas Drumetz, Bahram Ehsandoust, Jocelyn Chanussot, Bertrand Rivet, Massoud
  Babaie-Zadeh, and Christian Jutten,
\newblock ``Relationships between nonlinear and space-variant linear models in
  hyperspectral image unmixing,''
\newblock {\em IEEE Signal Processing Letters}, vol. 24, no. 10, pp.
  1567--1571, 2017.

\bibitem{Borsoi_2018_Fusion}
R.~A. {Borsoi}, T.~{Imbiriba}, and J.~C. {Moreira Bermudez},
\newblock ``{Super-Resolution for Hyperspectral and Multispectral Image Fusion
  Accounting for Seasonal Spectral Variability},''
\newblock {\em ArXiv e-prints}, Aug. 2018.

\bibitem{Borsoi2017_multiscale}
R.~A. {Borsoi}, T.~{Imbiriba}, J.~C. {Moreira Bermudez}, and C.~{Richard},
\newblock ``{A Fast Multiscale Spatial Regularization for Sparse Hyperspectral
  Unmixing},''
\newblock {\em IEEE Geoscience and Remote Sensing Letters}, 2018.

\bibitem{Borsoi_multiscaleVar_2018}
R.~A. {Borsoi}, T.~{Imbiriba}, and J.~C. {Moreira Bermudez},
\newblock ``{A Data Dependent Multiscale Model for Hyperspectral Unmixing With
  Spectral Variability},''
\newblock {\em ArXiv e-prints}, Aug. 2018.

\bibitem{qian2017tensorNMFunmixing}
Y.~Qian, F.~Xiong, S.~Zeng, J.~Zhou, and Y.~Y. Tang,
\newblock ``Matrix-vector nonnegative tensor factorization for blind unmixing
  of hyperspectral imagery,''
\newblock {\em IEEE Transactions on Geoscience and Remote Sensing}, vol. 55,
  no. 3, pp. 1776--1792, 2017.

\bibitem{imbiriba2018ULTRA}
T.~Imbiriba, R.~A. Borsoi, and J.~C.~M. Bermudez,
\newblock ``A low-rank tensor regularization strategy for hyperspectral
  unmixing,''
\newblock in {\em 2018 IEEE Statistical Signal Processing Workshop (SSP)},
  2018, pp. 373--377.

\bibitem{sidiropoulos2017tensor}
N.~D. Sidiropoulos, L.~De~Lathauwer, X.~Fu, K.~Huang, E.~E. Papalexakis, and
  C.~Faloutsos,
\newblock ``Tensor decomposition for signal processing and machine learning,''
\newblock {\em IEEE Transactions on Signal Processing}, vol. 65, no. 13, pp.
  3551--3582, 2017.

\bibitem{Boyd_admm_2011}
S.~Boyd, N.~Parikh, E.~Chu, B.~Peleato, and J.~Eckstein,
\newblock ``Distributed optimization and statistical learning via the
  alternating direction method of multipliers,''
\newblock {\em Found. Trends Mach. Learn.}, vol. 3, no. 1, pp. 1--122, Jan.
  2011.

\bibitem{Nascimento2005}
J.~M.~P. Nascimento and J.~M. Bioucas-Dias,
\newblock ``{Vertex Component Analysis}: A fast algorithm to unmix
  hyperspectral data,''
\newblock {\em IEEE Transactions on Geoscience and Remote Sensing}, vol. 43,
  no. 4, pp. 898--910, April 2005.

\bibitem{clark2003imaging}
R.~N. Clark, G.~A. Swayze, K.~E. Livo, R.~F. Kokaly, S.~J. Sutley, J.~B.
  Dalton, R.~R. McDougal, and C.~A. Gent,
\newblock ``Imaging spectroscopy: Earth and planetary remote sensing with the
  {USGS} tetracorder and expert systems,''
\newblock {\em Journal of Geophysical Research: Planets}, vol. 108, no. E12,
  2003.

\end{thebibliography}

\end{document}